*Article*

# Enhancing Cognitive Workload Classification Using Integrated LSTM Layers and CNNs for fNIRS Data Analysis

**Mehshan Ahmed Khan [1,*], Houshyar Asadi [1], Mohammad Reza Chalak Qazani [2], Adetokunbo Arogbonlo [1], Siamak Pedrammehr [1], Adnan Anwar [1], Hailing Zhou [3], Lei Wei [1], Asim Bhatti [1], Sam Oladazimi [4], Burhan Khan [1] and Saeid Nahavandi [5]**

[1] Institute for Intelligent Systems Research and Innovation, Deakin University, 75 Pigdons Rd, Waurn Ponds, VIC, 3216, Geelong, Australia; mehshan.khan@deakin.edu.au (M.A); houshyar.asadi@deakin.edu.au (H.A.); a.arogbonlo@deakin.edu.au (A.A.); s.pedrammehr@deakin.edu.au (S.P.); adnan.anwar@deakin.edu.au (A.A.); lei.wei@deakin.edu.au (L.W.); asim.bhatti@deakin.edu.au (A.B.); burhan.khan@deakin.edu.au (B.K.)
[2] Faculty of Computing and Information Technology (FoCIT), Sohar University, Sohar, Oman; m.r.chalakqazani@gmail.com
[3] School of Engineering, Swinburne University of Technology, Hawthorn, VIC 3122, Melbourne, Australia ; hailingzhou@swin.edu.au (H.Z.)
[4] Hermann-Hesse Str. 17 72221 Haiterbach, Germany ; sam.ozi@lenus-healthcare.de (S.O.)
[5] Swinburne Research, Swinburne University of Technology, Hawthorn, VIC 3122, Melbourne, Australia ; snahavandi@swin.edu.au (S.N.)
* Correspondence: mehshan.khan@deakin.edu.au; houshyar.asadi@deakin.edu.au (H.A.)

**Abstract:** Functional near-infrared spectroscopy (fNIRS) is employed as a non-invasive method to monitor functional brain activation by capturing changes in the concentrations of oxygenated hemoglobin (HbO) and deoxygenated hemoglobin (HbR). Various machine learning classification techniques have been utilized to distinguish cognitive states. However, conventional machine learning methods, although simpler to implement, undergo a complex pre-processing phase before network training and demonstrate reduced accuracy due to inadequate data preprocessing. Additionally, previous research in cognitive load assessment using fNIRS has predominantly focused on differentiating between two levels of mental workload. These studies mainly aim to classify low and high levels of cognitive load or distinguish between easy and difficult tasks. To address these limitations associated with conventional methods, this paper conducts a comprehensive exploration of the impact of Long Short-Term Memory (LSTM) layers on the effectiveness of Convolutional Neural Networks (CNNs) within deep learning models. This is to address the issues related to spatial feature overfitting and the lack of temporal dependencies in CNNs discussed in the previous studies. By integrating LSTM layers, the model can capture temporal dependencies in the fNIRS data, allowing for a more comprehensive understanding of cognitive states. The primary objective is to assess how incorporating LSTM layers enhances the performance of CNNs. The experimental results presented in this paper demonstrate that the integration of LSTM layers with convolutional layers results in an increase in the accuracy of deep learning models from 97.40% to 97.92%.

## 1. Introduction

Cognitive load, a fundamental concept in cognitive psychology, pertains to the quantity of information that the working memory can retain [1,2]. As the human brain possesses finite capacities for processing information simultaneously, delving into the intricacies of cognitive load provides valuable insights into the mechanisms that govern optimal learning and cognitive functioning. The n-back [3] task is a typical paradigm designed to administer a specific and measurable cognitive workload to individuals. In this task, participants are tasked with a dynamic challenge, requiring them to respond, typically by pressing a button, whenever the current stimulus matches with the stimulus presented n times earlier in the sequence. This cognitive challenge not only demands participants' sustained attention but also necessitates the effective engagement of their working memory.

In the field of neurophysiological research, functional near-infrared spectroscopy (fNIRS) has emerged as a non-invasive and adaptable neuroimaging technique, contributing significantly to the exploration of cognitive workload [4,5]. By capturing real-time changes in cerebral blood flow and oxygenation levels, fNIRS provides researchers with dynamic indicators of cognitive engagement. Its flexibility is particularly noteworthy as it enables studies in realistic and natural environments, adding ecological validity to the research. In comparison to conventional methods like electroencephalography (EEG) and functional Magnetic Resonance Imaging (fMRI), fNIRS demonstrates superior resistance to motion artifacts and environmental noise, making it particularly advantageous for research conducted in dynamic and ecologically valid settings [6].

fNIRS employs near-infrared light to assess the concentration and oxygenation levels of blood in tissue at depths ranging from 1 to 3 cm [7]. This technique relies on the emission of light within the near-infrared spectrum (650–900 nm) into the scalp, primarily the forehead region. The emitted light undergoes diffuse reflection as it traverses through the scalp, skull, and cortical areas of the brain. At these wavelengths, oxygenated hemoglobin (HbO) and deoxygenated hemoglobin (HbR) serve as the predominant absorbers, enabling precise measurements of their respective concentrations. The working principle of fNIRS involves the detection of a small fraction of the light initially emitted into the head that returns to the detector on the fNIRS probe. By analyzing the intensity of the returned light, researchers can calculate both the oxygenation level and the total blood volume in the observed tissue. This capability is rooted in the biological phenomenon known as neurovascular coupling: when a specific brain region becomes active, it experiences an increased demand for energy, triggering an uptick in blood flow to that region [8]. This hemodynamic response typically results in elevated concentrations of oxygenated hemoglobin and reduced levels of deoxygenated hemoglobin in the active region, as illustrated in Figure 1. The ability of fNIRS to measure localized brain activity with high specificity arises from its capacity to monitor these hemodynamic changes.

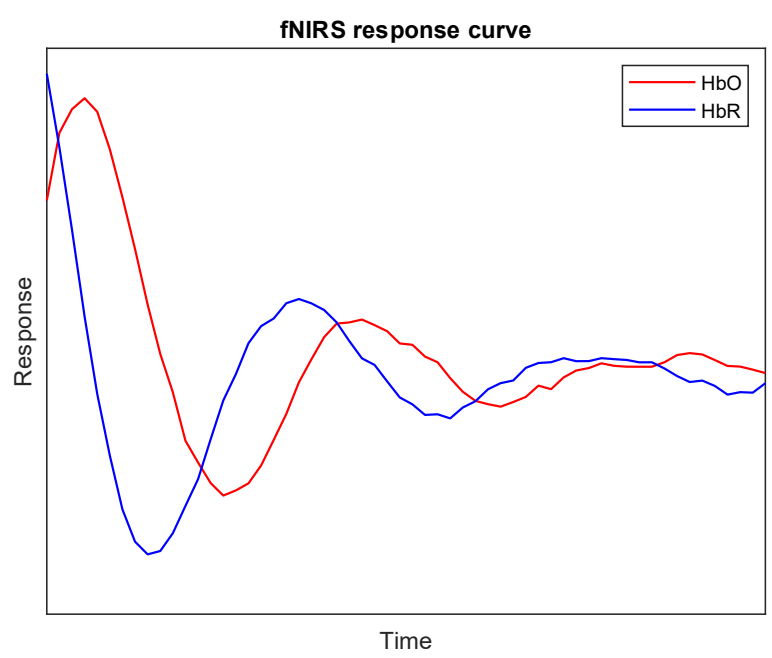

**Figure 1.** Visualization of hemodynamic changes in brain activity.

Calculations of hemoglobin concentration changes are made possible through the application of the modified Beer–Lambert Law [9]. This law provides a mathematical framework to transform the raw optical density data into quantifiable changes in HbO and HbR concentrations, enabling researchers to draw meaningful inferences about brain function and activity. In addition to its technical robustness, fNIRS offers distinct advantages in real-world applications due to its non-invasive nature, portability, and resistance to motion artifacts. These characteristics make it an invaluable tool for studying brain activity under naturalistic and dynamic conditions, expanding its applicability beyond traditional laboratory settings.

Past studies have employed several approaches to evaluate cognitive load through the analysis of fNIRS data. Conventionally, statistical tests have been employed on fNIRS signals for cognitive load assessment [10]. This involves subjecting fNIRS signals to statistical analyses, with a focus on the concentration of relative oxygenated hemoglobin (HbO). The significance of activation is commonly established through per-channel t-tests [11,12], providing insights into specific brain regions influenced by cognitive load. Moreover, researchers frequently used Analysis of Variance (ANOVA) tests, encompassing both one-way and two-way ANOVA, to facilitate group-level comparisons of mean activation [13,14].

Recognizing the challenges posed by the ever-expanding scale and complexity of fNIRS data, researchers are increasingly turning to machine learning (ML) and deep learning (DL) methodologies as promising alternatives. Due to the accessibility of cost-effective processing power, the utilization of ML or DL in diverse domains like image processing [15], signal processing [16], and remote sensing [17] has garnered considerable attention from researchers. In the field of neuroscience, constructing an ML model involves several stages, including data preprocessing, feature engineering, the development of ML or DL models, and a subsequent evaluation of performance.

Various studies have successfully employed a range of ML algorithms, such as Random Forest [18,19], k-Nearest Neighbors (k-NN) [20], Naive Bayes [21], Linear Discriminant Analysis (LDA) [22], and Support Vector Machines (SVM) [23], on fNIRS signals. However, challenges exist in the extraction and selection of features from these signals. The ML methods commonly employed for these tasks often suffer from limitations, particularly regarding scalability and efficiency [24]. One notable limitation lies in the process of extracting and selecting features, wherein traditional ML methods tend to fall short. These limitations become especially apparent when considering the computational cost, which escalates quadratically with respect to the parameters under consideration. The implication is a substantial increase in resource requirements, demanding vast amounts of labeled training data for supervised learning applications.

In contrast to the traditional ML approaches, DL adopts a paradigm shift by employing a deep neural network to handle the entire process. Specifically, autoencoders [25] and Convolutional Neural Networks (CNN) [26] have become important actors, emphasizing the resolution of optimization problems related to automatic feature extraction [27] and classification tasks for fNIRS signals [28]. The application of DL techniques, with their inherent capacity for automatic feature extraction and hierarchical representation learning, stands out as a noteworthy advancement in the field of fNIRS signal analysis. Therefore, this study aims to employ DL methodologies for the purpose of classifying mental work into distinct levels. In contrast to conventional machine learning approaches, which often rely on manual feature engineering, the focus here is on leveraging the capabilities of deep neural networks. Numerous deep learning models have been explored in the existing literature. However, our architecture outperforms the current state-of-the-art by introducing an innovative framework that utilizes the inherent capabilities of deep neural networks for the classification of mental work. One of the distinguishing features of our proposed model is its ability to automatically extract relevant features from fNIRS signals, which is an essential component in understanding the complex patterns indicative of different mental work levels.

Most of the studies conducted so far in the domain of fNIRS have focused on classifying two levels of workload using deep learning techniques [29–31]. While these studies provide valuable insights into cognitive load classification, it is noteworthy to highlight that previous studies conducted on the Tufts fNIRS dataset have predominantly concentrated on the differentiation between two levels of mental workload. This study addresses this limitation by utilizing the open-access Tufts fNIRS dataset [32], which consists of four distinct workload levels, providing a richer and more challenging dataset for cognitive load classification. Despite the availability of the Tufts dataset, there has been a noticeable lack of research leveraging this resource, especially in the context of lightweight deep learning models. The existing studies either do not explore this dataset extensively or rely on complex, computationally intensive architectures, which may not be suitable for real-time or resource-constrained applications. This study bridges this gap by employing a lightweight CNN-LSTM-based model designed to efficiently classify cognitive load across four workload levels. The hybrid architecture combines the spatial feature extraction capabilities of CNNs with the temporal dependency modeling strength of Long Short-Term Memory networks (LSTMs), making it both robust and computationally efficient. While CNN layers specialize in spatial feature extraction [33], LSTM layers excel in capturing temporal dependencies [34]. However, one significant challenge in comparing the performance of the proposed model with the previous studies lies in the lack of clarity regarding data preprocessing methods. The absence of standardized or well-documented preprocessing pipelines in prior research makes it difficult to replicate their methodologies and draw direct comparisons. Despite this challenge, the study demonstrates the effectiveness of the lightweight CNN-LSTM model, highlighting its potential for advancing workload classification using fNIRS data.

The primary objective of this study is to delve into the classification of mental workload by extending the scope beyond the conventional binary categorization. Specifically, this study centers its focus on distinguishing and classifying four distinct levels of mental workload by analyzing the publicly available Tufts fNIRS dataset [32]. To achieve this, we propose the implementation of a hybrid DL-based architecture that utilizes the capabilities of CNN in with LSTM. This fusion of CNN and LSTM layers presents a unique approach to fNIRS signal analysis [35]. The main contributions of this article can be summarized as follows: (1) A novel framework is introduced for classifying cognitive load levels using fNIRS signals, offering a structured approach to tackle multilevel cognitive load classification challenges. (2) The proposed framework leverages a hybrid CNN-LSTM-based model, combining the strengths of Convolutional Neural Networks (CNNs) for spatial feature extraction and Long Short-Term Memory (LSTM) networks for capturing temporal dependencies in fNIRS data. (3) The study demonstrates that fNIRS signals can be effectively utilized for automatic cognitive load classification. (4) The performance of the proposed framework is compared with traditional machine learning methods. Additionally, the impact of removing LSTM layers from the model is investigated to evaluate their contribution to classification accuracy. By integrating these components, our model provides a comprehensive solution for the challenges posed by fNIRS data, outperforming traditional models that often focus on either spatial or temporal aspects in isolation.

The remainder of this paper is as follows. Section 2 will describe the suggested CNN and LSTM architecture designed for fNIRS measurements. Following this, Section 3 will provide an overview of the experiment, results, and validation, with the subsequent discussion also presented within the same section. The study concludes in Section 4.

## 2. Related Work

Physiological measurement techniques such as electroencephalography (EEG), fNIRS, Electrocardiography, Electromyography, and eye-tracking are extensively used to identify and analyze cognitive load [36–38]. These methods enable the recording of physiological signals in real time during task performance, offering researchers powerful tools to quantify and assess cognitive load under varying conditions. Among these, fNIRS have been particularly favored in cognitive load research due to their ability to provide meaningful insights into brain activity while accommodating more ecologically valid experimental settings, which often require participant mobility [25]. Cognitive load is commonly investigated using adaptations of the ISO-standardized n-back task, a widely accepted paradigm for inducing and measuring cognitive workload. The n-back task's design challenges participants' working memory and attention, making it a suitable method for eliciting varying levels of cognitive load.

Before the rise in deep neural networks as the dominant approach for feature extraction, earlier research on fNIRS data heavily relied on hand-crafted feature engineering. As fNIRS data are inherently time-series in nature, researchers often segmented the data into specific time windows and calculated statistical features such as the slope, standard deviation, skewness, and kurtosis of the HbO and HbR signals. Additionally, metrics like the zero-lagged correlation between HbO and HbR were used to capture interdependencies between the two hemodynamic signals. Aghajani et al. [39] demonstrated this approach by classifying cognitive load levels induced by the n-back task. They extracted statistical features from HbO and HbR signals, focusing on their sensitivity to cognitive load changes. By employing an SVM combined with a moving window method, they achieved a mean accuracy of 74.8% for binary classification. Similarly, Herff et al. [40] used an LDA to classify mental tasks, achieving accuracies of 71%, 70%, and 62% for mental arithmetic, word generation, and mental rotation tasks, respectively, against a resting baseline. In another study, Liu et al. [41] focused on average HbO and HbR amplitude changes as features for classifying cognitive load levels in n-back tasks. Their work achieved a mean accuracy of 53.9% using LDA. Hong et al. [42] also applied LDA to distinguish between mental arithmetic and left- and right-hand motor imagery tasks, achieving an average classification accuracy of 75.6%. Shin et al. [43] used SVMs to classify mental arithmetic tasks against a baseline, with performances of around 77% accuracy for tasks performed with eyes open and 75% with eyes closed. Beyond SVM and LDA, other standard machine learning models like k-NN were also explored. For example, k-NN was employed to classify three different mental workload levels on n-back tasks, achieving accuracies of up to 52.08% [44]. In addition to statistical features, regression techniques were employed to extract meaningful information from fNIRS data. Herff et al. [45] used linear regression to fit a slope to the data within specific time windows during n-back tasks. This approach yielded accuracies of 81%, 80%, and 72% for classifying 3-back, 2-back, and 1-back tasks against a relaxed state, respectively. While these methods provided early insights into cognitive load and task classification, they were limited by their reliance on manually selected features. The performance of these models often depended on the quality of the feature engineering process and their ability to capture the nuanced and complex patterns inherent in fNIRS data. Moreover, the linear assumptions underlying many of these statistical and regression-based features often failed to account for non-linear relationships in the data.

CNNs, a remarkable branch of deep learning, have revolutionized image-driven pattern recognition tasks and extended their applications across diverse fields, including natural language processing, voice recognition, and brain–computer interfaces. In recent years, CNNs have been increasingly applied to neuroimaging data, including functional near-infrared spectroscopy (fNIRS) [46,47], to address complex challenges such as early diagnosis of neurological conditions and task classification. One of the critical applications of CNNs is in the early detection of mild cognitive impairment, a precursor to Alzheimer's disease (AD). Early intervention can play a significant role in managing and potentially mitigating the progression of AD. Zyrianov et al. [48] demonstrated that CNN-based deep learning models applied to fNIRS data could effectively distinguish MCI patients from healthy controls. The study involved participants performing cognitive tasks such as the n-back task, Stroop test, and verbal fluency tasks, with the CNN achieving notable accuracy in classification. Further advancing this area, B. Lakshmi Priya et al. [49] combined the Empirical Wavelet Transform with wavelet scattering coefficients and fed this enriched feature set into a recurrent neural network (RNN). This approach analyzed brain wave patterns in response to thought processes and visual stimuli, resulting in improved classification performance for cognitive impairments. Processing raw fNIRS time-series data for classification presents significant challenges due to the complex and high-dimensional nature of the data. Wang et al. [50] proposed converting fNIRS time-series data into images to improve classification accuracy. This transformation enables CNNs to leverage their inherent strengths in image recognition to identify patterns in fNIRS data. Trakoolwilaiwan et al. [51] utilized four distinct CNN architectures to extract features from fNIRS signals. The results showed that CNNs outperformed traditional machine learning models, such as SVMs and artificial neural networks (ANNs), which relied on hand-crafted features (e.g., mean, variance, kurtosis, skewness, peak, and slope of HbO and HbR signals). The CNN approach demonstrated superior accuracy, highlighting its potential for automated feature extraction in fNIRS data analysis. Saadati et al. [52] employed CNNs combined with a moving window method to classify cognitive load levels from fNIRS data. Their approach achieved an average classification accuracy of 82%, significantly improving upon traditional methods. This highlights CNNs' ability to capture the temporal and spatial complexities of fNIRS signals, making them a valuable tool for cognitive load research. Innovative techniques have further expanded CNN applications in fNIRS and time-series data. Wickramaratne et al. [53] utilized Gramian Angular Summation Fields to convert time-series data into image-like representations. These images were then processed using CNNs to classify tasks such as mental arithmetic, motor imagery, and idle states. The CNN model was benchmarked against methods like sLDA, ANN, and Bi-LSTM. While sLDA is often preferred for multi-class linear classification problems, Bi-LSTM (Bidirectional Long Short-Term Memory) networks leverage both past and future information, enhancing their performance for sequential data. Despite the strengths of these alternative methods, CNNs consistently demonstrated superior accuracy and robustness in handling complex fNIRS datasets.

Conventional machine learning models, such as Support Vector Machines (SVMs), Linear Discriminant Analysis (LDA), and k-Nearest Neighbors (kNN), also exhibit limitations in the context of cognitive load classification. SVMs, while effective for binary or small multi-class problems, struggle with high-dimensional datasets like raw fNIRS signals and rely on hand-crafted features, which limits their scalability [54,55]. Similarly, LDA assumes linear separability between classes, which is rarely realistic for the non-linear and high-dimensional nature of fNIRS data, leading to suboptimal performance for tasks involving multiple levels of cognitive load. kNN, although conceptually simple, suffers from computational inefficiency for large datasets and lacks the capacity to learn internal representations, making it unsuitable for complex tasks [56]. Artificial neural networks (ANNs) are versatile machine learning models capable of modeling complex relationships in data. However, while ANNs are effective in various applications, their performance often relies on extensive feature engineering and require more computational power, making them less suitable for raw or structured data [57,58]. In contrast, CNNs are specifically designed to handle structured data and perform automated feature extraction. Studies have consistently demonstrated that CNNs outperform traditional ANNs, primarily due to their robustness in capturing spatial patterns directly from raw or pre-processed data [59,60]. Recurrent neural network (RNN) variants, such as Bidirectional Long Short-Term Memory networks (Bi-LSTM), have shown promise for sequential data analysis by capturing temporal dependencies. However, Bi-LSTMs alone lack the ability to extract spatial features effectively, which is critical for analyzing brain activity patterns in fNIRS data. Consequently, standalone Bi-LSTM models may underperform in scenarios requiring simultaneous spatial and temporal feature extraction [61].

To address these limitations, we propose the use of a hybrid CNN-LSTM model for classifying cognitive load across four levels. This approach combines the spatial feature extraction capabilities of CNNs with the temporal dependency modeling strength of LSTMs [61,62]. CNNs effectively process fNIRS data by extracting spatial patterns from either raw signals or image-like representations of the data. LSTMs complement this by capturing temporal dynamics, allowing the model to account for the evolving nature of cognitive load over time. By leveraging the strengths of CNNs and LSTMs, the hybrid model not only improves classification accuracy but also enhances scalability and robustness to noise and variability in fNIRS signals.

## 3. Methodology

In our study, we introduce a robust 1-dimensional Convolutional Neural Network (1D-CNN) coupled with LSTM layers to formulate a predictive model for mental workload based on fNIRS data. Figure 2 illustrates the flow chart of the proposed CNN and LSTM-based architecture. This design incorporates baseline convolutional blocks to extract features from the input data. Subsequently, the extracted features undergo additional processing through LSTM layers to capture temporal dependencies in the data, thereby improving the accuracy of the classification output.

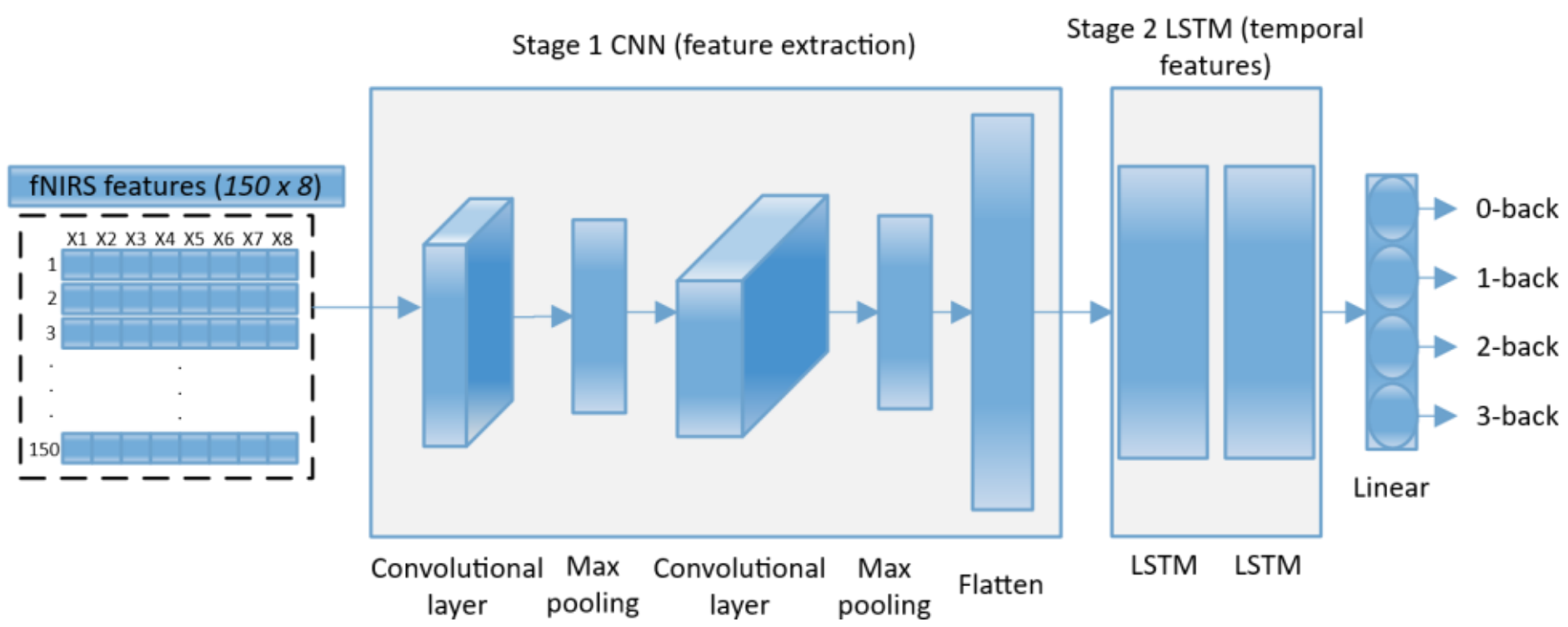

**Figure 2.** The diagram illustrates the proposed architecture, incorporating both CNN and LSTM layers.

### *3.1. Convolution Neural Networks (CNN)s*

In recent years, CNNs have emerged as a groundbreaking paradigm in the field of artificial intelligence. A key component of CNNs is the convolutional kernel's weight-sharing technique for the convolutional kernel [63]. One of the distinctive strengths of CNNs is its remarkable ability to learn implicit features without the need for intricate pre-processing or a distinct feature extraction process [64]. The baseline blocks of our proposed model consist of the convolutional layer, Rectified Linear Unit (Relu) activation layer, and max pooling layer. The convolutional operation, as expressed by Equation (1), encapsulates the fundamental process by which a given input tensor $X$ interacts with a set of learnable filters $W$ and bias $b$.

$$Y[i] = f\left(\sum_{j=1}^{k} X[i+j-1].W[j] + b\right) \quad (1)$$

Following the convolution operation, the activation function plays a crucial role by nonlinearly transforming the resulting output values. This transformation is instrumental in mapping the original multi-dimensional features, contributing to an augmented linear separability of the extracted features. In our model, we employed the Rectified Linear Unit (Relu) activation function, denoted by Equation (2), to introduce a non-linearity that enhances the model's ability to capture complex patterns.

$$Y[i] = \max(0, X[i]) \quad (2)$$

The maximum pooling layer serves the dual purpose of diminishing the dimensionality of the convolved and extracted features while also contributing to the generalization capabilities of the network. The mathematical representation of the max pooling operation for a given input $X$, with a pooling size denoted as $p$, is formally defined in Equation (3). This operation involves selecting the maximum value from each local region of size $p$, effectively downsizing the spatial dimensions of the features.

$$Y[i] = \max(i.p:(i+1).p) \quad (3)$$

The LSTM [17] network extends the capabilities of the recurrent neural network (RNN) [65]. While RNNs utilize a directed cycle structure, transferring the output of a hidden layer to the same hidden layer, LSTM architectures excel in capturing and learning from the inherent temporal dynamics in data [66]. Unlike traditional RNNs, LSTMs are designed to tackle issues like vanishing and exploding gradients [67], commonly hindering the effective learning of long-term dependencies in sequential data. This is accomplished through specialized memory cells and gating mechanisms, allowing LSTMs to selectively retain and discard information over extended sequences. The distinctive feature of LSTMs lies in their ability to memorize information over extended time intervals. This is particularly relevant for fNIRS data as it exhibits dependencies on previous data points. LSTM possess internal memory units that allow them to retain and utilize information from previous time steps, making them suitable for modeling cognitive load dynamics. The equations governing the behavior of these models are as follows:

$$i_t = \sigma(U_i * X_t + V_i * H_{t-1} + Z_i \circ C_{t-1} + b_i), \quad (4)$$

$$f_t = \sigma\left(U_f * X_t + V_f * H_{t-1} + Z_f \circ C_{t-1} + b_f\right), \quad (5)$$

$$o_t = \sigma\left(U_o * X_t + V_f * H_{t-1} + Z_f \circ C_{t-1} + b_f\right), \quad (6)$$

$$\widetilde{C_t} = \tanh(U_c * X_t + V_c * H_{t-1} + b_c) \quad (7)$$

$$C_t = f_t \circ C_{t-1} + i_t \circ \widetilde{C_t} \quad (8)$$

$$H_t = o_t * \tanh(C_t) \quad (9)$$

In Equations (4)–(9), '$*$' represents convolution, and '$\circ$' represents the Hadamard product. Cell states are denoted as $C_1, \ldots, C_t$ and hidden states as $H_t$, $i_t$, $f_t$, $o_t$ denote the input gate, forget gate, and output gate, respectively. $\sigma$ denotes the sigmoid activation function. $U_i$, $U_f$, $U_o$, $V_i$, $V_f$, $V_o$, $V_c$, $Z_i$, $Z_f$, $Z_o$ are 2D convolution kernels. $b_i$, $b_c$, $b_f$ and $b_o$ are the bias terms.

### 3.2. CNN and LSTM-Based Proposed Model

The neural network architecture outlined in Figure 2 is tailored for the classification of cognitive load using functional near-infrared spectroscopy (fNIRS) data. In the fNIRS, where the data often involves both spatial and temporal aspects, the combination of 1D-CNN and LSTM layers offers a robust approach. Table 1 denotes the length of the input sequence as "L" and batch size as "BS". The CNN components, including convolutional layers, Relu activation functions, and max pooling layers, enable the model to extract spatial features and patterns from fNIRS. Convolutional layers apply filters to capture local patterns, Relu introduces non-linearity, and max pooling reduces spatial dimensions for computational efficiency. The Flatten layer (flatten) then transforms multi-dimensional feature maps into a one-dimensional vector for further processing.

**Table 1.** Hierarchical feature representation in the proposed architecture utilizing CNN and LSTM for cognitive load classification.

| Deep Neural Network Layers | Input Size | Output Size |
|---|---|---|
| Conv1d | (BS, 8, L) | (BS, 16, L) |
| Relu | (BS, 16, L) | (BS, 16, L) |
| MaxPool1d | (BS, 16, L) | (BS, 16, L/2) |
| Conv1d | (BS, 16, L/2) | (BS, 32, L/2) |
| Relu | (BS, 32, L/2) | (BS, 32, L/2) |
| MaxPool1d | (BS, 32, L/2) | (BS, 32, L/4) |
| Flatten | (BS, 32, L/4) | (BS, 32 * L/4) |
| LSTM | (BS, 32 * L/4) | (BS, 64) |
| LSTM | (BS, 64) | (BS, 64) |
| Linear | (BS, 64) | (BS, 128) |
| Relu | (BS, 128) | (BS, 128) |
| Linear | (BS, 128) | (BS, 4) |

On the other hand, the inclusion of LSTM layers in the architecture addresses the network's capability to handle sequential data, making it well suited for fNIRS data. LSTMs possess a gating mechanism that selectively updates and forgets information over sequences, allowing the model to capture temporal patterns effectively. The choice of two LSTM layers suggests a focus on learning intricate dependencies in sequential data. The fully connected layers that follow the LSTM layers perform the final classification based on the features learned from both the CNN and LSTM layers.

## 4. Experiment

In the assessment of the proposed CNN and LSTM-based models designed for the classification of mental workload, we employed the Tufts fNIRS open-access dataset [32]. This dataset encompasses fNIRS data derived from a cohort of 68 participants engaged in a series of controlled n-back tasks, encompassing 0-back, 1-back, 2-back, and 3-back scenarios. The raw fNIRS measurements comprise temporal traces of alternating current intensity and changes in phase at two distinct wavelengths (690 and 830 nm), employing a modulation frequency of 110 MHz. To mitigate the influence of respiration, heartbeat, and drift artifacts, each univariate time series underwent bandpass filtering using a 3rd-order zero-phase Butterworth filter, with a retention range of 0.001–0.2 Hz.

### 4.1. Preprocessing

The segmentation of fNIRS signals involved the use of overlapping windows, each spanning 30 s with a stride of 0.6 s. This approach was adopted based on the recommendations of the dataset authors who observed that 30 s windows achieved the highest accuracy for individual subjects. At each timestep within these windows, eight numerical measurements were recorded, resulting in the creation of eight features.

In the context of the 30 s window size, each feature vector comprised 150 fNIRS measurements, contributing to a length of $150 \times 8$. The adoption of the $150 \times 8$ feature size was deemed essential not only to facilitate the real-time implementation of deep learning models but also to enable these models to capture and learn temporal dependencies present within the data frames. We employed the Scikit-learn library (version 1.5.2) to handle the data splitting process, ensuring reproducibility by setting the random state parameter to 10. This parameter guarantees that the random splitting of the dataset into training and testing subsets remains consistent across multiple runs of the experiment. Specifically, the dataset was randomly divided into two subsets: 80% of the data were allocated for training purposes, while the remaining 20% was reserved for testing. This approach ensures a fair distribution of samples for evaluating the model's performance. The total number of samples in the dataset amounted to 101,184. After splitting, the training subset contained 80,947 samples, and the testing subset comprised 20,237 samples. This distribution was deemed suitable for effectively training and evaluating the model. Given the small input size of $150 \times 8$, a batch size of 1024 was chosen to facilitate faster convergence during training. A larger batch size reduces the number of iterations required per epoch, thereby accelerating the training process without compromising model accuracy. The model was trained over 1000 epochs to ensure sufficient exposure to the data for optimizing its performance. This number of epochs was selected after considering the model's capacity, the dataset size, and the computational resources available, balancing the need for thorough training with efficiency.

### *4.2. Results and Discussion*

The employed methodology is implemented using Python in conjunction with the PyTorch [68] framework. The computational infrastructure utilized for model training and evaluation is a custom-built machine equipped with an Intel Core i9 12900K CPU, 32 GB of RAM, and a dedicated NVIDIA RTX 3090ti GPU.

A CNN can be conceptualized as a parameterized function, and its overall performance is intricately tied to the selection of optimal parameters. To train a deep learning model, the Adam [69] optimizer is employed, with a specified learning rate of 0.001. Additionally, the cross-entropy [70] loss function has been incorporated into the model architecture. Our training spanning a total of 1000 epochs as shown in Figure 3.

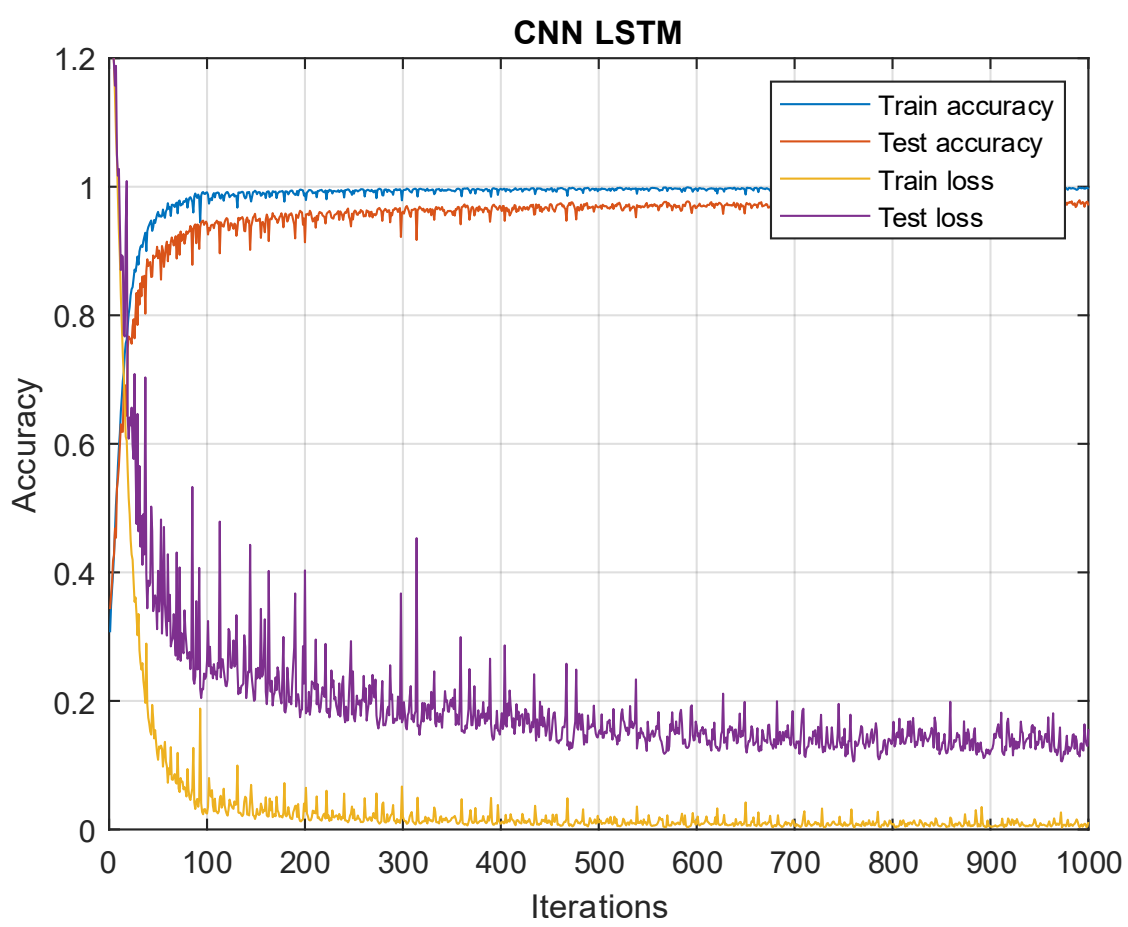

**Figure 3.** Training and testing accuracies in a proposed CNN and LSTM-based model.

Throughout the 1000 epochs of training, we closely monitored the performance metrics of the model, specifically observing the dynamic interplay between accuracy and loss. The upward trajectory of accuracy signifies an improvement in the model's ability to correctly classify samples, demonstrating its capacity to learn and generalize effectively. On the other hand, the downward trend in loss reflects a reduction in the disparity between the predicted and actual distributions, underscoring the model's proficiency in minimizing classification errors. It signifies that the model has reached a state where its predictions align closely with ground truth labels and further training may yield diminishing returns. This convergence not only validates the effectiveness of the chosen parameters, including the Adam optimizer and the cross-entropy loss function, but also underscores the model's capacity to capture intricate patterns within the data. The confusion matrix for the proposed CNN and LSTM-based models, as presented in Figure 4, reveals a notably low incidence of misclassifications. The model demonstrates a high level of precision in correctly identifying and categorizing samples with only a minimal number of instances where it deviates from the ground truth labels.

Convolutional Neural Networks (CNNs) have demonstrated a powerful ability to extract and learn meaningful features from data with minimal preprocessing, as outlined in the preceding discussion. However, a significant challenge with deep learning models, including CNNs, is the risk of overfitting, particularly when the dataset size is limited. Overfitting occurs when a model learns to perform exceptionally well on training data but fails to generalize to unseen data, leading to poor performance on validation or test sets. In Figure 3, the learning rate, though relatively small, facilitates smooth convergence within 1000 epochs, achieving minimal loss. Similarly, Figure 4 highlights the model's highly accurate classification results, reflecting the strength of the proposed architecture. Nevertheless, small sample sizes exacerbate the risk of overfitting, as the model may memorize specific patterns in the training data rather than learning generalized representations. Although dropout layers have not been implemented in the proposed model, overfitting appears to have been mitigated through other means inherent to the architecture and training process. Specifically, the use of max pooling layers plays a significant role in addressing overfitting. By reducing the spatial dimensions of the feature maps, max pooling limits the model's capacity to overfit to local, noise-induced patterns in the data. This operation forces the network to focus on the most prominent features within each pooling region, effectively acting as a form of regularization. Furthermore, the relatively small learning rate likely contributed to a gradual and steady optimization process, preventing the model from converging too quickly to suboptimal solutions that could lead to overfitting. Additionally, achieving convergence within 1000 epochs suggests that the model was carefully monitored during training to avoid excessive training beyond the point of generalization. Techniques such as early stopping, though not explicitly stated, may have been employed to halt training once validation performance plateaued or began to degrade. Finally, the architecture's design, including the combination of convolutional layers and max pooling, inherently controls the model's complexity by reducing dimensionality and focusing on high-level features. This design choice ensures that the model is neither excessively complex nor overly simplistic, enabling it to generalize effectively while minimizing the risk of overfitting.

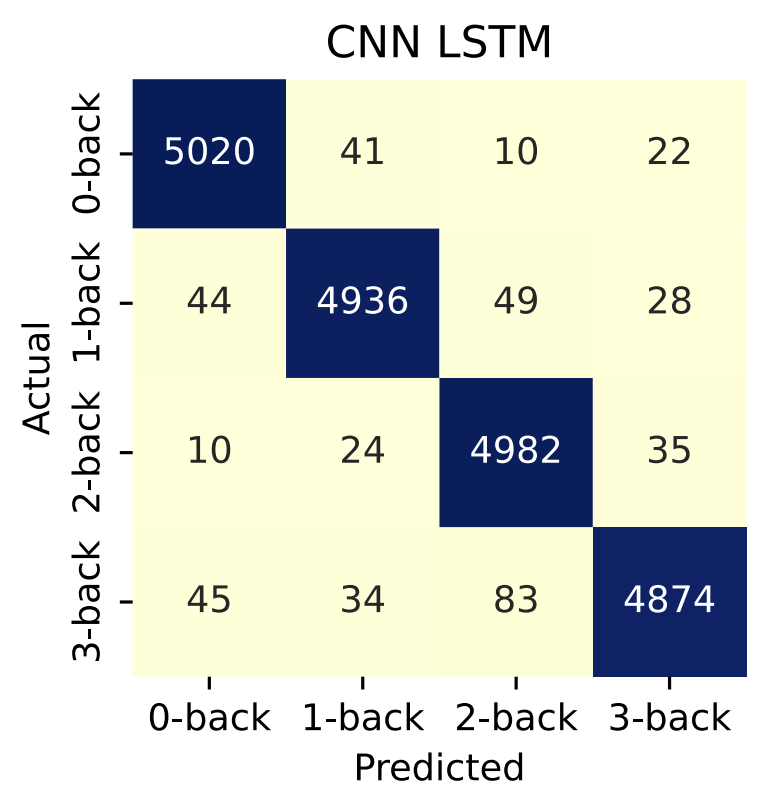

**Figure 4.** Confusion matrix of proposed CNN- and LSTM-based model.

To validate the efficacy of our proposed model, we experimented by systematically removing all LSTM layers from the architecture and retraining the model on the identical dataset. The results revealed that the modified model, lacking the LSTM layers, exhibited a marginally lower level of accuracy compared to the original model.

Upon analyzing the training curve illustrated in Figure 5, it became evident that although the model did converge, there were discernible fluctuations present. The fluctuations in accuracy and loss can be attributed to the inability of the CNN-only model to effectively capture sequential relationships inherent in the dataset. CNNs are primarily designed for spatial feature extraction, and when applied to time-series data without a mechanism to account for sequential order, they may struggle to model temporal correlations. This limitation results in inconsistent learning patterns, as evidenced by the oscillations in the training curve. Additionally, the CNN-only model lacks the ability to retain context from earlier time steps, which is crucial for datasets where temporal dependencies play a significant role in distinguishing patterns across time. This gap in contextual understanding leads to the model being overly reliant on immediate, localized patterns rather than learning long-term dependencies. Furthermore, the absence of LSTM layers led to the model being more sensitive to noise and local minima during the optimization process, exacerbating instability in convergence. This sensitivity arises because CNNs, without recurrent mechanisms, lack the capacity to smooth out noisy or incomplete temporal patterns, causing the model to overfit to minor variations in the data rather than focusing on the broader trends. Moreover, the lack of temporal feature integration means that the CNN-only model struggles with capturing transitions or temporal dynamics between data points, which are often critical for accurate classification. These limitations not only hinder the model's ability to converge smoothly but also result in reduced generalization, as reflected by the higher variance in accuracy and loss across epochs. A closer examination of the confusion matrix for the CNN model in Figure 6, which lacks LSTM layers, reveals a higher frequency of misclassifications compared to the proposed CNN LSTM-based model.

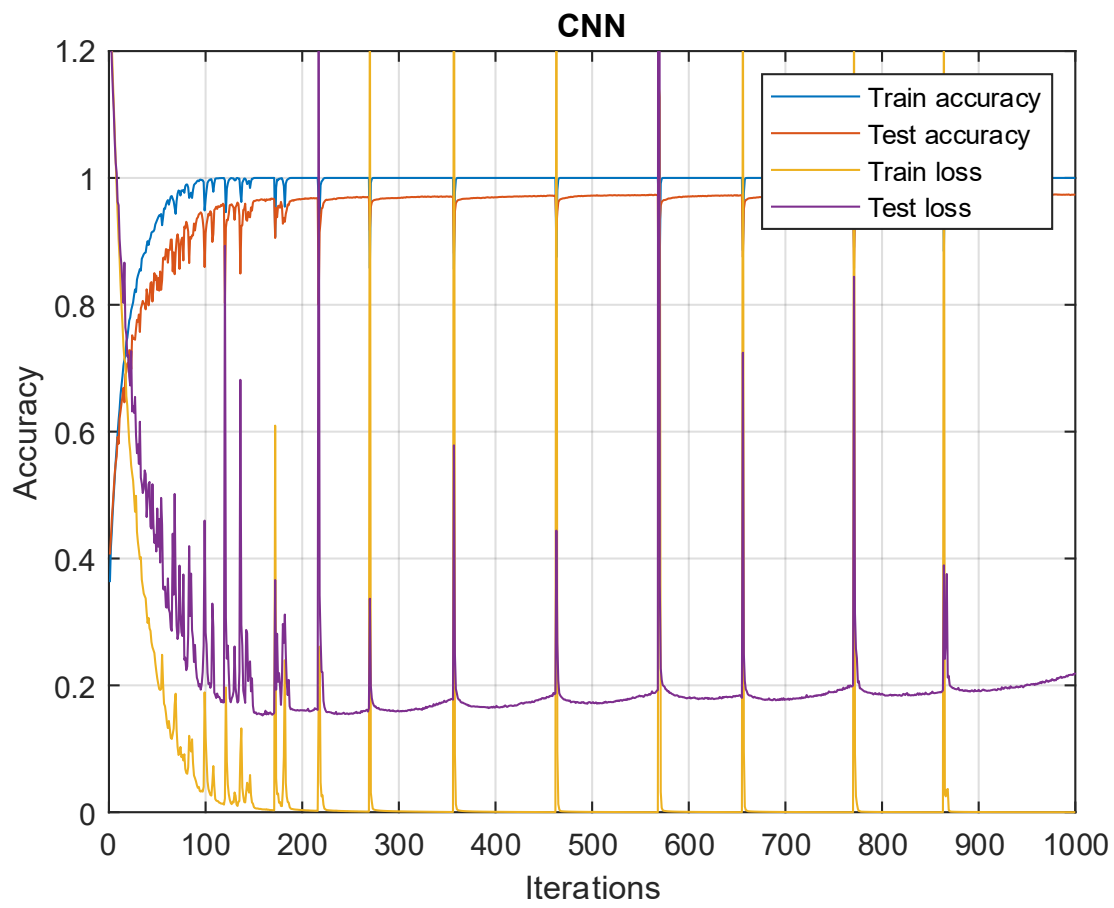

**Figure 5.** Training and testing accuracies in a CNN-based model.

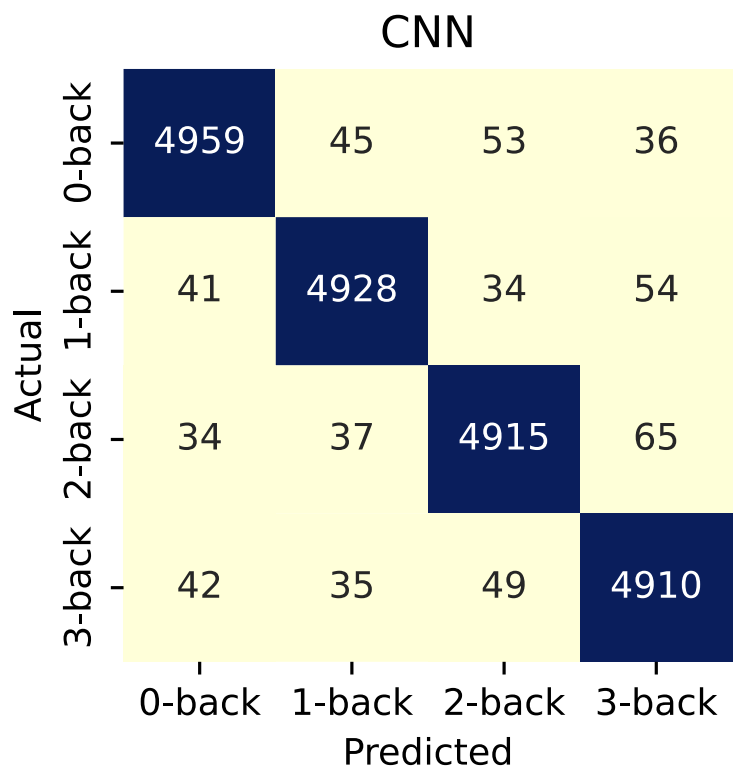

**Figure 6.** Confusion matrix of CNN model.

To assess the performance of the proposed CNN and LSTM-based models, a comparison has been made against traditional ML models [19]. The performance evaluation is encapsulated in Figure 7, which presents the confusion matrix for a range of ML algorithms. In contrast to the remarkable performance observed in the CNN and LSTM models, the traditional ML models, including Naive Bayes and Nearest Centroid, exhibit a notable disparity in their classification accuracy. Figure 7 illustrates that these models suffer from a higher incidence of misclassifications, signifying a limited capacity to discern and categorize samples accurately. Among the ML algorithms, Decision Trees and k-NN emerge as more robust performers, showcasing comparatively better accuracy and a reduced number of misclassifications. Interestingly, the performance of Decision Trees aligns closely with the proposed DL-based model, emphasizing the effectiveness of both approaches in handling the classification task. However, it is noteworthy that the Decision Tree model exhibits a slightly higher rate of misclassifications when compared to the DL counterparts.

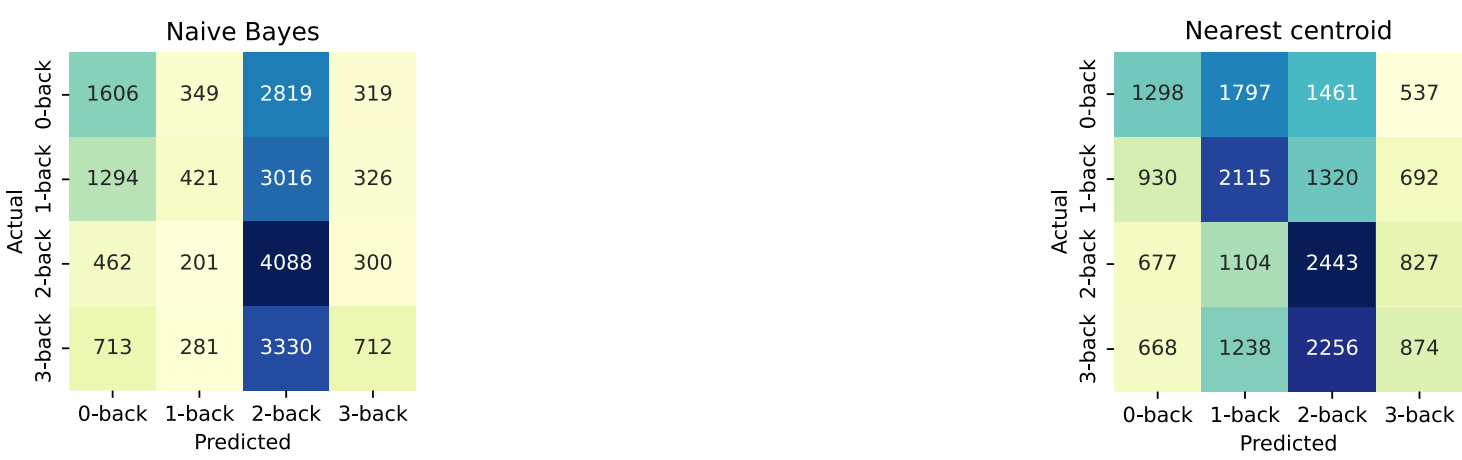

(**a**) (**b**)

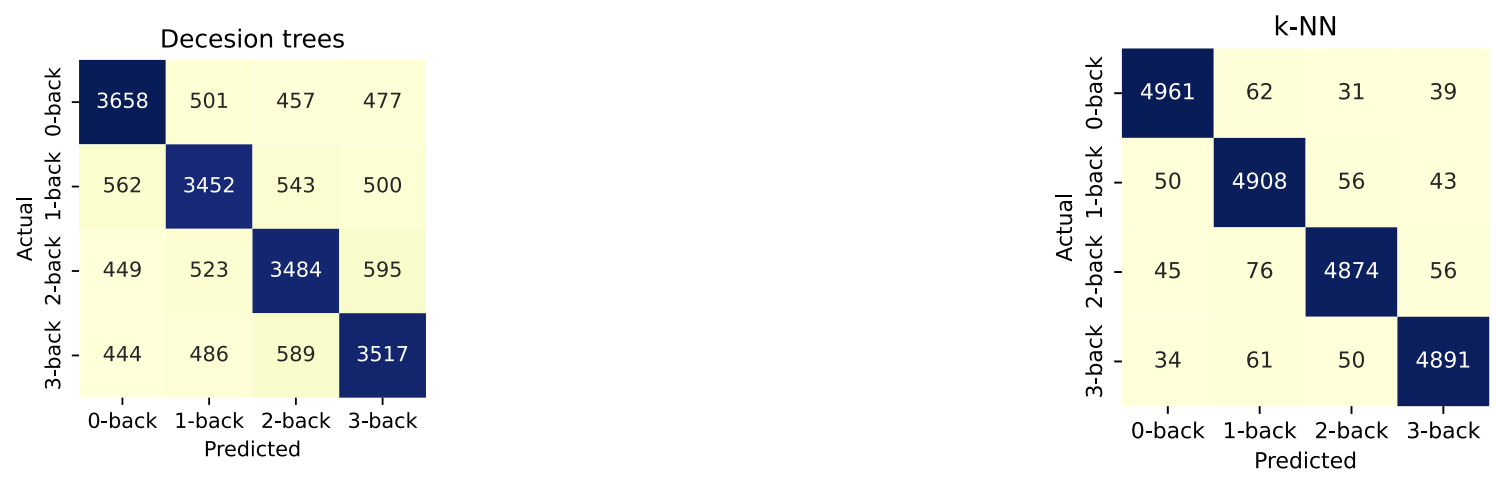

**Figure 7.** Confusion matrix of various machine learning classifiers (**a**) Naïve bayes, (**b**) Nearest Centroid, (**c**) Decision Trees, and (**d**) k-NN.

In addition to comparing the proposed CNN and LSTM models with traditional ML models, Table 2 presents a comprehensive evaluation of their performance using various classifiers and key metrics. The table offers an in-depth overview of each classifier's performance, encompassing accuracy, F1-score, AUC (Area Under the Curve), precision, and recall. Notably, Naïve Bayes and Nearest Centroid exhibit the lowest accuracy and F1-score percentages, with Naïve Bayes demonstrating slightly higher precision. Decision Trees consistently perform well, achieving 69.72% across accuracy, F1-score, AUC, precision, and recall. Among ML classifiers, k-NN outperforms other ML classifiers and achieves 97.02% across accuracy, F1-score, and balanced precision and recall. The proposed CNN with LSTM layers outperforms both traditional ML models and a CNN without LSTM layers across all metrics, showcasing its robustness in handling the classification task.

**Table 2.** Assessing the effectiveness of classification algorithms with precision, recall, F1-score, and AUC metrics.

| | Evaluation Metrics | | | | |
|---|---|---|---|---|---|
| **Classifiers** | **Accuracy** | **F1-Score** | **AUC** | **Precision** | **Recall** |
| Naïve Bayes | 0.3373 | 0.2859 | 0.5712 | 0.3671 | 0.3373 |
| Nearest Centroid | 0.3325 | 0.3207 | 0.5803 | 0.3316 | 0.3325 |
| Decision Trees | 0.6972 | 0.6972 | 0.6105 | 0.6972 | 0.6972 |
| k-NN | 0.9702 | 0.9702 | 0.6628 | 0.9702 | 0.9702 |
| CNN without LSTM Layers | 0.9740 | 0.9740 | 0.6638 | 0.9740 | 0.9740 |
| Proposed CNN LSTM | **0.9789** | **0.9789** | **0.6704** | **0.9790** | **0.9789** |

The superior performance of the proposed CNN and LSTM-based models, especially when compared to traditional ML algorithms, underscores the transformative potential of deep learning for the evaluation cognitive load assessment using fNIRS signals. As highlighted in Figure 7 and Table 2, the CNN and LSTM models exhibit a notable advantage in accuracy, F1-score, AUC, precision, and recall, offering a comprehensive and detailed evaluation of cognitive load levels.

Table 3 presents the time taken by the CNN and CNN-LSTM models for training in minutes. The training times were measured on a standard hardware setup using the respective architectures. As shown, the CNN model completed training in approximately 7.79 min, while the CNN-LSTM model required slightly more time, approximately 9.76 min. This increase in training time for the CNN-LSTM model is expected due to the additional complexity introduced by the LSTM layers, which capture temporal dependencies in the data. These results highlight the trade-off between the additional computational cost and the improved ability of the CNN-LSTM model to handle temporal features, which can lead to enhanced model performance in tasks that require sequential data understanding. Table 4 presents a comprehensive comparison of the inference times for CNN and CNN-LSTM models across GPU and CPU platforms, offering valuable insights into their computational performance. The results underscore the inherent trade-off between computational complexity and model performance, particularly when integrating LSTM layers into the architecture to enhance its capability for capturing temporal dependencies. To ensure the reliability and robustness of the recorded results, a meticulous evaluation process was employed. Initially, each model was executed 1000 times as a warm-up phase. This preliminary step was crucial for excluding anomalous results that may arise due to factors such as initial data transfer overheads, including the time required to move data from system RAM to the GPU cache. These transfer processes can introduce slight deviations in inference time, particularly during the initial runs, which might not accurately reflect steady-state performance. After the warm-up phase, the models were executed 100,000 times for each configuration to compute the average inference time. This approach provided a statistically robust foundation for comparison, ensuring that the reported times reflect consistent and realistic performance. By basing the average inference time on a large number of runs, the evaluation minimizes the impact of random fluctuations and transient inefficiencies, delivering a more precise measurement of the models' computational efficiency under realistic usage scenarios.

**Table 3.** Training time.

| Models | Time (Minutes) |
|---|---|
| CNN | 7.79 |
| CNN-LSTM | 9.76 |

The CNN model demonstrates consistently lower inference times on both GPU and CPU, primarily due to its streamlined structure that relies on convolutional layers for feature extraction. On the GPU, the average inference time for the CNN is 0.0891 ms, significantly faster than the 0.2153 ms observed for the CNN-LSTM model. Similarly, on the CPU, the CNN achieves an average inference time of 0.21535 ms compared to the 0.83560 ms required by the CNN-LSTM. The slightly higher inference times for the CNN-LSTM model can be attributed to the inclusion of two LSTM layers. These layers enhance the model's ability to capture temporal dependencies in the data but also introduce additional computational overhead. The complexity of LSTM operations, especially during sequential data processing, contributes to the observed increase in inference times. This analysis demonstrates the balance between model simplicity and the need for more sophisticated architectures to handle temporal patterns. While CNN models are ideal for scenarios requiring rapid inference, the CNN-LSTM architecture offers a compelling alternative for tasks where temporal modeling is critical, albeit at a slightly higher computational cost. By running the models multiple times and computing average values over 100,000 iterations, this study ensures that the reported inference times are accurate, consistent, and representative of real-world usage scenarios.

**Table 4.** Inference time metrics on GPU and CPU for CNN and CNN-LSTM architectures.

| Metric | GPU: CNN Inference Time (ms) | GPU: CNN-LSTM Inference Time (ms) | CPU: CNN Inference Time (ms) | CPU: CNN-LSTM Inference Time (ms) |
|---|---|---|---|---|
| Average Time | 0.0891 ± 0.0266 | 0.2153 ± 0.0413 | 0.21535 ± 0.04132 | 0.83560 ± 0.1463 |
| Minimum Time | 0.0317 | 0.087 | 0.087 | 0.643 |
| Maximum Time | 0.7516 | 1.4888 | 1.4888 | 2.4185 |

Deep learning's capability for automatic feature extraction and hierarchical representation learning proves especially beneficial in fNIRS-based cognitive load assessment. In contrast to traditional ML models relying on manually engineered features, the CNN and LSTM models autonomously discern intricate spatial and temporal patterns within the fNIRS data. This inherent ability leads to a more precise and robust classification of mental workload levels, crucial for comprehending cognitive processes. Additionally, these models outperform traditional ML counterparts like Naive Bayes and Nearest Centroid and even compete effectively with Decision Trees and k-NN. The automated and adaptive nature of deep learning models allows them to adjust to the dynamic and intricate nature of cognitive processes, providing a more flexible and scalable solution for real-world applications. The data collection process for this study was conducted ensuring high-quality input data to train the model effectively. However, the potential for label noise in the dataset remains a consideration. Label noise can stem from subtle inconsistencies in annotating cognitive load levels or variability in participants' responses during data collection [71]. Despite these challenges, the proposed CNN-LSTM-based framework is designed to be inherently robust to such noise. The CNN component excels at extracting meaningful spatial features from fNIRS data, while the LSTM layers effectively capture temporal dependencies, enabling the model to perform well even in the presence of imperfectly labeled data. The lightweight architecture of the model provides an additional advantage, achieving minimal inference times measured in milliseconds. This characteristic makes the framework particularly suitable for real-time applications, such as adaptive learning systems, real-time workload monitoring, and human–computer interaction scenarios, where rapid and reliable classification of cognitive load is essential. Furthermore, the model's low computational requirements make it deployable on resource-constrained devices, such as wearables or edge computing systems, broadening its applicability to practical use cases. Although the dataset used in this study is publicly available and centralized for model training, future implementations of this framework could benefit from decentralized data collection and aggregation [72]. In real-world scenarios, cognitive load data may be collected from diverse sources, such as multiple devices or participants, under varying conditions. By adopting a centralized approach to aggregate such data, the model's performance could be further enhanced through fine-tuning with a more comprehensive dataset. Centralization would also facilitate the identification and mitigation of any inconsistencies or label noise present in the data. This study's contributions not only demonstrate the effectiveness of the proposed CNN-LSTM-based model but also emphasize its potential for scalability and adaptability in future applications. With the ability to leverage publicly available datasets and adapt to decentralized data collection methods, the framework is well positioned to advance multi-level cognitive load classification using fNIRS data in real-world, dynamic settings.

## 5. Conclusions

The article investigates the incorporation of LSTM layers into CNNs. Unlike many neuroscience studies that predominantly rely on a singular model, our research adopts a hybrid approach. Specifically, we utilized fNIRS signal data sourced from the open access TUFTS datasets in our investigation. Our goal is to classify cognitive states, specifically 0-back, 1-back, 2-back, and 3-back, thereby expanding the scope beyond the conventional binary workload classification. Our work extends upon previous studies that predominantly concentrated on employing deep learning models to classify only two levels of workload using fNIRS data. This endeavor aims not only to contribute to a more comprehensive understanding of cognitive states but also enhances the applicability of our findings to real-world scenarios where cognitive demands vary across a continuum. One of the main focuses of our research is to evaluate the impact of integrating LSTM layers into CNNs within the framework of deep learning models. By utilizing the strengths of both convolutional and LSTM operators, we can effectively capture spatial and temporal dependencies, leading to enhanced performance.

Our experimental findings provide compelling evidence of the effectiveness of integrating LSTM layers into the CNN architecture. The model's accuracy demonstrated a noteworthy improvement, increasing from 97.40% to 97.82%. This improvement can be attributed to the LSTM layers facilitating the model in capturing and leveraging temporal dependencies within the fNIRS data. Furthermore, our study involved a comparative analysis with traditional ML methods. Among the various ML classifiers employed, k-NN outperformed other ML classifiers. It is noteworthy that while DL methods employed in our research showcased superior performance compared to ML classifiers, it is essential to acknowledge the inherent trade-off. Designing and optimizing DL models necessitates a substantial investment in time and effort. In addition, we compared the inference time of the CNN and CNN-LSTM models. Adding the LSTM layers did not significantly increase the inference time, demonstrating that the hybrid CNN-LSTM model can achieve higher performance without a substantial computational cost. In consideration of the potential for further improvement, it appears that extending the depth of the CNN and LSTM beyond the current proposal could yield enhanced results. In our future endeavors, we plan to advance the scope of our research by proposing a more robust deep CNN model. This advanced model will feature an increased number of layers and optimal parameterization, allowing it to unveil more intricate non-linear structures progressively.

**Author Contributions:** Conceptualization, M.A.K. and H.A.; data curation, M.A.K.; formal analysis, M.A.K.; methodology, M.A.K. and H.A.; project administration, H.A.; supervision, H.A.; visualization, M.A.K., M.R.C.Q., A.A. (Adetokunbo Arogbonlo), S.P., A.A. (Adnan Anwar), A.B., S.N., S.O., H.Z., L.W. and B.K.; validation, M.A.K., H.A., H.Z. and L.W.; writing—original draft, M.A.K.; writing—review and editing, H.A., M.R.C.Q., A.A. (Adetokunbo Arogbonlo), S.P., A.A. (Adnan Anwar), A.B., S.N., S.O., H.A., H.Z. and B.K. All authors have read and agreed to the published version of the manuscript.

**Funding:** This research received no external funding.

**Data Availability Statement:** We used the publicly available Tufts dataset for our experiment, which can be accessed at the following link: https://tufts-hci-lab.github.io/code_and_datasets/fNIRS2MW.html. The dataset was accessed on January 1st, 2023. Additionally, the source code for the deep learning models implemented and used in this article is available for public access on GitHub. You can find the repository at: https://github.com/mehshankhan/fNIRS_tufts_CNN_lSTM?tab=readme-ov-file.

**Acknowledgments:** This research was supported by the Australian Research Council (ARC) (Project ID: DE210101623).

**Conflicts of Interest:** The authors declare no conflicts of interest.

## References

1. Yin, Y.; Chen, S.; Song, T.; Zhou, Q.; Shao, Y. Cognitive Load Moderates the Effects of Total Sleep Deprivation on Working Memory: Evidence from Event-Related Potentials. *Brain Sci.* **2023**, *13*, 898.
2. Farkish, A.; Bosaghzadeh, A.; Amiri, S.H.; Ebrahimpour, R. Evaluating the Effects of Educational Multimedia Design Principles on Cognitive Load Using EEG Signal Analysis. *Educ. Inf. Technol.* **2022**, *28*, 2827–2843.
3. Kirchner, W.K. Age differences in short-term retention of rapidly changing information. *J. Exp. Psychol.* **1958**, *55*, 352.
4. Hirshfield, L.M.; Wickens, C.; Doherty, E.; Spencer, C.; Williams, T.; Hayne, L. Toward Workload-Based Adaptive Automation: The Utility of fNIRS for Measuring Load in Multiple Resources in the Brain. *Int. J. Hum.–Comput. Interact.* **2024**, *40*, 7404–7430.
5. Flanagan, K.; Saikia, M.J. Consumer-Grade Electroencephalogram and Functional Near-Infrared Spectroscopy Neurofeedback Technologies for Mental Health and Wellbeing. *Sensors* **2023**, *23*, 8482.

6. Zhuang, C.; Meidenbauer, K.L.; Kardan, O.; Stier, A.J.; Choe, K.W.; Cardenas-Iniguez, C.; Huppert, T.J.; Berman, M.G. Scale Invariance in fNIRS as a Measurement of Cognitive Load. *Cortex* **2022**, *154*, 62–76.
7. Villringer, A.; Chance, B. Non-invasive optical spectroscopy and imaging of human brain function. *Trends Neurosci.* **1997**, *20*, 435–442.
8. D'Esposito, M.; Zarahn, E.; Aguirre, G.K. Event-related functional MRI: Implications for cognitive psychology. *Psychol. Bull.* **1999**, *125*, 155.
9. Kocsis, L.; Herman, P.; Eke, A. The modified Beer–Lambert law revisited. *Phys. Med. Biol.* **2006**, *51*, N91.
10. Cakar, S.; Yavuz, F.G. Nested and robust modeling techniques for fNIRS data with demographics and experiment related factors in n-back task. *Neurosci. Res.* **2023**, *186*, 59–72.
11. Broadbent, D.P.; D'Innocenzo, G.; Ellmers, T.J.; Parsler, J.; Szameitat, A.J.; Bishop, D.T. Cognitive load, working memory capacity and driving performance: A preliminary fNIRS and eye tracking study. *Transp. Res. Part F Traffic Psychol. Behav.* **2023**, *92*, 121–132.
12. Polat, M.D.; Izzetoglu, K.; Aksoy, M.E.; Kitapcioglu, D.; Usseli, T.; Yoner, S.I. Cognitive Load Quantified via Functional Near Infrared Spectroscopy During Immersive Training with VR Based Basic Life Support Learning Modules in Hostile Environment. In Proceedings of the International Conference on Human-Computer Interaction, Copenhagen, Denmark, 23–28 July 2023; Springer: Cham, Switzerland, 2023.
13. Ayman, S.U.; Arrafuzzaman, A.; Rahman, M.A. Subject Dependent Cognitive Load Level Classification from fNIRS Signal Using Support Vector Machine. In Proceedings of the International Conference on Information and Communication Technology for Development: ICICTD 2022, Khulna, Bangladesh, 29–30 July 2022; Springer: Singapore, 2023.
14. Han, Y.; Huang, J.; Yin, Y.; Chen, H. From Brain to Worksite: The Role of fNIRS in Cognitive Studies and Worker Safety. *Front. Public Health* **2023**, *11*, 1256895.
15. Khan, M.A.; Khan, M.A.; Ahmed, F.; Mittal, M.; Goyal, L.M.; Hemanth, D.J.; Satapathy, S.C. Gastrointestinal diseases segmentation and classification based on duo-deep architectures. *Pattern Recognit. Lett.* **2020**, *131*, 193–204.
16. Ghandorh, H.; Khan, M.Z.; Alsufyani, R.; Khan, M.; Alsofayan, Y.M.; Khan, A.A.; Alahmari, A.A. An ICU Admission Predictive Model for COVID-19 Patients in Saudi Arabia. *Int. J. Adv. Comput. Sci. Appl.* **2021**, *12*, 553–567.
17. Boulila, W.; Ghandorh, H.; Khan, M.A.; Ahmed, F.; Ahmad, J. A novel CNN-LSTM-based approach to predict urban expansion. *Ecol. Inform.* **2021**, *64*, 101325.
18. Oku, A.Y.A.; Sato, J.R. Predicting student performance using machine learning in fNIRS data. *Front. Hum. Neurosci.* **2021**, *15*, 622224.
19. Khan, M.A.; Asadi, H.; Hoang, T.; Lim, C.P.; Nahavandi, S. Measuring Cognitive Load: Leveraging fNIRS and Machine Learning for Classification of Workload Levels. In Proceedings of the International Conference on Neural Information Processing, Changsha, China, 20–23 November 2023; Springer: Singapore, 2023.
20. Fernandez Rojas, R.; Huang, X.; Ou, K.-L. A machine learning approach for the identification of a biomarker of human pain using fNIRS. *Sci. Rep.* **2019**, *9*, 5645.
21. Hasan, M.Z.; Islam, S.M.R. Suitibility Investigation of the different classifiers in fNIRS signal classification. In Proceedings of the 2020 IEEE Region 10 Symposium (TENSYMP), Dhaka, Bangladesh, 5–7 June 2020; IEEE: Piscataway, NJ, USA, 2020.
22. Gemignani, J. Classification of fNIRS data with LDA and SVM: A proof-of-concept for application in infant studies. In Proceedings of the 2021 43rd Annual International Conference of the IEEE Engineering in Medicine & Biology Society (EMBC), Mexico City, Mexico, 1–5 November 2021; IEEE: Piscataway, NJ, USA, 2021.
23. Chen, L.; Li, Q.; Song, H.; Gao, R.; Yang, J.; Dong, W.; Dang, W. Classification of schizophrenia using general linear model and support vector machine via fNIRS. *Phys. Eng. Sci. Med.* **2020**, *43*, 1151–1160.
24. Ghandorh, H.; Khan, M.Z.; Khan, M.A.; Alsofayan, Y.M.; Alahmari, A.A.; Khan, A.A. Predicting ICU Admission for COVID-19 Patients in Saudi Arabia: A Comparative Study of AdaBoost and Bagging Methods. *Int. J. Adv. Comput. Sci. Appl.* **2024**, *15*, 1249–1266.
25. Liu, R.; Reimer, B.; Song, S.; Mehler, B.; Solovey, E. Unsupervised fNIRS feature extraction with CAE and ESN autoencoder for driver cognitive load classification. *J. Neural Eng.* **2021**, *18*, 036002.
26. Kumar, A.; Karmakar, S.; Agarwal, I.; Pal, T. Mental Workload Classification with One-Dimensional CNN Using fNIRS Signal. In Proceedings of the International Conference on Pattern Recognition and Machine Intelligence, Kolkata, India, 12–15 December 2023; Springer: Cham, Switzerland, 2023.
27. Zafar, A.; Dad Kallu, K.; Atif Yaqub, M.; Ali, M.U.; Hyuk Byun, J.; Yoon, M.; Su Kim, K. A Hybrid GCN and Filter-Based Framework for Channel and Feature Selection: An fNIRS-BCI Study. *Int. J. Intell. Syst.* **2023**, *2023*, 8812844.

28. Karmakar, S.; Kamilya, S.; Dey, P.; Guhathakurta, P.K.; Dalui, M.; Bera, T.K.; Halder, S.; Koley, C.; Pal, T.; Basu, A. Real time detection of cognitive load using fNIRS: A deep learning approach. *Biomed. Signal Process. Control.* **2023**, *80*, 104227.
29. Cakar, S.; Yavuz, F.G. Enhancing Mental Workload Classification in Neuroscience: Deep Learning Approaches. *Res. Sq.* 2024, *preprint*.
30. Jing, Y.; Wang, W.; Wang, J.; Jiao, Y.; Xiang, K.; Lin, T.; Shi, W.; Hou, Z.G. Transformer Based Cross-Subject Mental Workload Classification Using FNIRS for Real-World Application. In Proceedings of the 2023 45th Annual International Conference of the IEEE Engineering in Medicine & Biology Society (EMBC), Sydney, Australia, 24–27 July 2023; IEEE: Piscataway, NJ, USA, 2023.
31. Lin, J.; Chung, F.-L.; Wang, S. Reservoir Transfer Learning Through Source Enhancement Reconstructed with Reservoir Stein Discrepancy. *IEEE Internet Things J.* **2024**, *11*, 22803–22815.
32. Huang, Z.; Wang, L.; Blaney, G.; Slaughter, C.; McKeon, D.; Zhou, Z.; Jacob, R.; Hughes, M.C. The Tufts fNIRS mental workload dataset & benchmark for brain-computer interfaces that generalize. In Proceedings of the Thirty-Fifth Conference on Neural Information Processing Systems Datasets and Benchmarks Track (Round 2), 2021.
33. Yu, C.; Han, R.; Song, M.; Liu, C.; Chang, C.-I. A simplified 2D-3D CNN architecture for hyperspectral image classification based on spatial–spectral fusion. *IEEE J. Sel. Top. Appl. Earth Obs. Remote Sens.* **2020**, *13*, 2485–2501.
34. Honnashamaiah, A.; Rathnakara, S. ECG signal classification using CNN & LSTM with Aquila Optimization technique. *Tuijin Jishu/J. Propuls. Technol.* **2023**, *44*, 5075–5087.
35. Khan, M.A.; Asadi, H.; Qazani, M.R.C.; Lim, C.P.; Nahavandi, S. Functional near-infrared spectroscopy (fNIRS) and Eye tracking for Cognitive Load classification in a Driving Simulator Using Deep Learning. *arXiv* **2024**, arXiv:2408.06349.
36. Sweller, J. Cognitive load during problem solving: Effects on learning. *Cogn. Sci.* **1988**, *12*, 257–285.
37. Ahmad, M.I.; Keller, I.; Robb, D.A.; Lohan, K.S. A framework to estimate cognitive load using physiological data. *Pers. Ubiquitous Comput.* **2023**, *27*, 2027–2041.
38. Srinivas, N.S.; Vimalan, V.; Padmanabhan, P.; Gulyás, B. An overview on cognitive function enhancement through physical exercises. *Brain Sci.* **2021**, *11*, 1289.
39. Aghajani, H.; Garbey, M.; Omurtag, A. Measuring Mental Workload with EEG+fNIRS. *Front. Hum. Neurosci.* **2017**, *11*, 359.
40. Herff, C.; Heger, D.; Putze, F.; Hennrich, J.; Fortmann, O.; Schultz, T. Classification of mental tasks in the prefrontal cortex using fNIRS. In Proceedings of the 2013 35th Annual International Conference of the IEEE Engineering in Medicine and Biology Society (EMBC), Osaka, Japan, 3–7 July 2013; IEEE: Piscataway, NJ, USA, 2013.
41. Liu, Y.; Ayaz, H.; Shewokis, P.A. Multisubject "learning" for mental workload classification using concurrent EEG, fNIRS, and physiological measures. *Front. Hum. Neurosci.* **2017**, *11*, 389.
42. Hong, K.-S.; Naseer, N.; Kim, Y.-H. Classification of prefrontal and motor cortex signals for three-class fNIRS–BCI. *Neurosci. Lett.* **2015**, *587*, 87–92.
43. Shin, J.; Müller, K.-R.; Hwang, H.-J. Near-infrared spectroscopy (NIRS)-based eyes-closed brain-computer interface (BCI) using prefrontal cortex activation due to mental arithmetic. *Sci. Rep.* **2016**, *6*, 36203.
44. Kesedžić, I.; Šarlija, M.; Božek, J.; Popović, S.; Ćosić, K. Classification of cognitive load based on neurophysiological features from functional near-infrared spectroscopy and electrocardiography signals on n-back task. *IEEE Sens. J.* **2020**, *21*, 14131–14140.
45. Herff, C.; Heger, D.; Fortmann, O.; Hennrich, J.; Putze, F.; Schultz, T. Mental workload during n-back task—Quantified in the prefrontal cortex using fNIRS. *Front. Hum. Neurosci.* **2014**, *7*, 935.
46. Khan, M.A.; Asadi, H.; Qazani, M.R.C.; Arogbonlo, A.; Nahavandi, S.; Lim, C.P. Predicting cognitive load in immersive driving scenarios with a hybrid CNN-RNN model. *arXiv* **2024**, arXiv:2408.06350.
47. Khan, M.A.; Asadi, H.; Zhang, L.; Qazani, M.R.C.; Oladazimi, S.; Kiong, L.C.; Lim, C.P.; Nahavandi, S. Application of artificial intelligence in cognitive load analysis using functional near-infrared spectroscopy: A systematic review. *Expert Syst. Appl.* **2024**, 123717.
48. Yang, D.; Huang, R.; Yoo, S.-H.; Shin, M.-J.; Yoon, J.A.; Shin, Y.-I.; Hong, K.-S. Detection of mild cognitive impairment using convolutional neural network: Temporal-feature maps of functional near-infrared spectroscopy. *Front. Aging Neurosci.* **2020**, *12*, 141.
49. Jayalakshmy, S.; Pragatheeswaran, J.K.; Saraswathi, D.; Poonguzhali, N. Scattering convolutional network based predictive model for cognitive activity of brain using empirical wavelet decomposition. *Biomed. Signal Process. Control* **2021**, *66*, 102501.
50. Wang, Z.; Oates, T. Imaging time-series to improve classification and imputation. *arXiv* **2015**, arXiv:1506.00327.
51. Trakoolwilaiwan, T.; Behboodi, B.; Lee, J.; Kim, K.; Choi, J.-W. Convolutional neural network for high-accuracy functional near-infrared spectroscopy in a brain–computer interface: Three-class classification of rest, right-, and left-hand motor execution. *Neurophotonics* **2018**, *5*, 011008–011008.

52. Saadati, M.; Nelson, J.; Ayaz, H. Convolutional neural network for hybrid fNIRS-EEG mental workload classification. In *Advances in Neuroergonomics and Cognitive Engineering: Proceedings of the AHFE 2019 International Conference on Neuroergonomics and Cognitive Engineering, and the AHFE International Conference on Industrial Cognitive Ergonomics and Engineering Psychology, Washington, DC, USA, 24–28 July 2019*; Springer: Cham, Switzerland, 2019.
53. Wickramaratne, S.D.; Mahmud, M.S. A deep learning based ternary task classification system using gramian angular summation field in fNIRS neuroimaging data. In Proceedings of the 2020 IEEE International Conference on E-health Networking, Application & Services (HEALTHCOM), Shenzhen, China, 1–2 March 2021; IEEE: Piscataway, NJ, USA, 2021.
54. Lakshmi, B.; Kumar, S.S.; Sai, N.R.; Kumar, K.P.V.; Kumar, G.S.C. Wlan intrusion detection system based on svm. In Proceedings of the 2022 International Conference on Sustainable Computing and Data Communication Systems (ICSCDS), Erode, India, 7–9 April 2022; IEEE: Piscataway, NJ, USA, 2022.
55. Wang, J.; Shi, Q. Short-term traffic speed forecasting hybrid model based on chaos–wavelet analysis-support vector machine theory. *Transp. Res. Part C Emerg. Technol.* **2013**, *27*, 219–232.
56. Zhang, S. Challenges in KNN classification. *IEEE Trans. Knowl. Data Eng.* **2021**, *34*, 4663–4675.
57. Alzubaidi, L.; Zhang, J.; Humaidi, A.J.; Al-Dujaili, A.; Duan, Y.; Al-Shamma, O.; Santamaría, J.; Fadhel, M.A.; Al-Amidie, M.; Farhan, L. Review of deep learning: Concepts, CNN architectures, challenges, applications, future directions. *J. Big Data* **2021**, *8*, 1–74.
58. Mirakowski, A. Comparison of the Effectiveness of ANN and CNN in Image Classification. In Proceedings of the International Conference on Computational Collective Intelligence, Leipzig, Germany, 9–11 September 2024; Springer: Cham, Switzerland, 2024.
59. Sharma, S.; Shivhare, S.N.; Singh, N.; Kumar, K. Computationally efficient ann model for small-scale problems. In *Machine Intelligence and Signal Analysis*; Springer: Singapore, 2019.
60. Hasan, M.; Ullah, S.; Khan, M.J.; Khurshid, K. Comparative analysis of SVM, ANN and CNN for classifying vegetation species using hyperspectral thermal infrared data. *Int. Arch. Photogramm. Remote Sens. Spat. Inf. Sci.* **2019**, *42*, 1861–1868.
61. Sahoo, A.R.; Chakraborty, P. Hybrid CNN Bi-LSTM neural network for Hyperspectral image classification. *arXiv* **2024**, arXiv:2402.10026.
62. Hassan, N.; Miah, A.S.M.; Shin, J. A Deep Bidirectional LSTM Model Enhanced by Transfer-Learning-Based Feature Extraction for Dynamic Human Activity Recognition. *Appl. Sci.* **2024**, *14*, 603.
63. Xie, S.; Liu, L.; Sun, G.; Pan, B.; Lang, L.; Guo, P. Enhanced E-commerce Fraud Prediction Based on a Convolutional Neural Network Model. *CMC-Comput. Mater. Contin.* **2023**, *75*, 1107–1117.
64. Elakiya, V.; Puviarasan, N.; Aruna, P. Detection of violence using mosaicking and DFE-WLSRF: Deep feature extraction with weighted least square with random forest. *Multimed. Tools Appl.* **2023**, *83*, 40873–40908.
65. Bhanja, S.; Das, A. Deep neural network for multivariate time-series forecasting. In Proceedings of the International Conference on Frontiers in Computing and Systems: COMSYS 2020, Jalpaiguri, India, 13–15 January 2020;Springer: Singapore, 2021.
66. Song, X.; Liu, Y.; Xue, L.; Wang, J.; Zhang, J.; Wang, J.; Jiang, L.; Cheng, Z. Time-series well performance prediction based on Long Short-Term Memory (LSTM) neural network model. *J. Pet. Sci. Eng.* **2020**, *186*, 106682.
67. Wang, Q.; Peng, R.-Q.; Wang, J.-Q.; Li, Z.; Qu, H.-B. NEWLSTM: An optimized long short-term memory language model for sequence prediction. *IEEE Access* **2020**, *8*, 65395–65401.
68. Paszke, A.; Gross, S.; Chintala, S.; Chanan, G.; Yang, E.; DeVito, Z.; Lin, Z.; Desmaison, A.; Antiga, L.; Lerer, A. Automatic differentiation in pytorch. 2017.
69. Kingma, D.P.; Ba, J. Adam: A method for stochastic optimization. *arXiv* **2014**, arXiv:1412.6980.
70. Chong, K.; Huang, Y.; Koo, C.; Sherif, M.; Ahmed, A.N.; El-Shafie, A. Investigation of cross-entropy-based streamflow forecasting through an efficient interpretable automated search process. *Appl. Water Sci.* **2023**, *13*, 6.
71. Yang, B.; Lei, Y.; Li, X.; Li, N.; Nandi, A.K. Label recovery and trajectory designable network for transfer fault diagnosis of machines with incorrect annotation. *IEEE/CAA J. Autom. Sin.* **2024**, *11*, 932–945.
72. Yang, B.; Lei, Y.; Li, X.; Li, N. Targeted transfer learning through distribution barycenter medium for intelligent fault diagnosis of machines with data decentralization. *Expert Syst. Appl.* **2024**, *244*, 122997.